# Deep Learning-Based Multi-Modal Fusion for Robust Robot Perception and Navigation

1st Delun Lai*
*School of Electrical Engineering and Telecommunications*
*University of New South Wales*
Sydney, Australia
delunlai928@gmail.com

2nd Yeyubei Zhang
*School of Engineering and Applied Science*
*University of Pennsylvania*
Philadelphia, USA
joycezh@alumni.upenn.edu

3rd Yunchong Liu
*School of Engineering and Applied Science*
*University of Pennsylvania*
Philadelphia, USA
yunchong@alumni.upenn.edu

4th Chaojie Li
*School of Electrical Engineering and Telecommunications*
*University of New South Wales*
Sydney, Australia
chaojie.li@unsw.edu.au

5th Huadong Mo
*School of Systems and Computing*
*University of New South Wales*
Canberra, Australia
huadong.mo@unsw.edu.au

*Abstract*—This paper introduces a novel deep learning-based multimodal fusion architecture aimed at enhancing the perception capabilities of autonomous navigation robots in complex environments. By utilizing innovative feature extraction modules, adaptive fusion strategies, and time-series modeling mechanisms, the system effectively integrates RGB images and LiDAR data. The key contributions of this work are as follows: a. the design of a lightweight feature extraction network to enhance feature representation; b. the development of an adaptive weighted cross-modal fusion strategy to improve system robustness; and c. the incorporation of time-series information modeling to boost dynamic scene perception accuracy. Experimental results on the KITTI dataset demonstrate that the proposed approach increases navigation and positioning accuracy by 3.5% and 2.2%, respectively, while maintaining real-time performance. This work provides a novel solution for autonomous robot navigation in complex environments.

## I. Introduction

With the rapid development of robotics, autonomous navigation capability has become a core requirement for robotic systems. In practical applications, robots need to accurately perceive and reliably navigate in dynamically changing environments. However, the complexity and uncertainty of the environment and various disturbing factors have brought great challenges to the perception system of robots [1].

Traditional single-modal perception methods can achieve good results under specific conditions but often perform poorly in complex environments [2]. For example, vision systems are susceptible to light variations and weather conditions, while LiDAR (Light Detection and Ranging) may produce noisy data under precipitating weather. Therefore, how to effectively fuse multiple sensor data to achieve robust environment perception has become a key issue in current research [3].

In recent years, the development of deep learning has provided new solutions for multimodal fusion. Existing deep learning methods can be mainly categorized into (1) early fusion: fusing multimodal data directly at the input layer; (2) feature-level fusion: fusing features of different modalities at the intermediate layer of the network; and (3) decision-level fusion: fusing the independent prediction results of each modality. However, these methods still have problems such as low efficiency of feature extraction, fixed fusion strategy, and neglect of timing information [4-8].

To address the above problems, this paper proposes a novel deep learning multimodal fusion architecture. The main contributions are as follows:

- A lightweight multimodal feature extraction module is designed to improve feature quality and reduce computational complexity through an improved attention mechanism [9,10].
- An adaptive cross-modal fusion strategy is proposed to dynamically adjust the fusion weight according to the reliability of different modal data to improve system robustness [10].
- A temporal modeling mechanism is introduced to effectively utilize the temporal dependency between consecutive frames to improve the perception accuracy in dynamic scenes [11].
- A large number of experiments are carried out on the KITTI dataset to verify the effectiveness of the proposed method [12].

## II. Method

### A. Problem Definition

In autonomous navigation tasks, our robot system simultaneously obtains two important sensor data: RGB images taken by the camera and point cloud data scanned by the LiDAR. Among them, RGB images provide rich visual information, including the color, texture and appearance characteristics of the environment. In contrast, the LiDAR point cloud data provides accurate three-dimensional spatial information, including the position, shape and distance information of the object [13-15].

The core goal of this paper is to design an intelligent data processing system that can process these two types of data simultaneously and output the following two key results:

- A comprehensive understanding of the current environment [16].
- Corresponding navigation instructions [17].

In the process of achieving this goal, we face three significant challenges: First, we need to process two completely different sets of data effectively. Camera images are two-dimensional, with neatly arranged pixels, while LiDAR data is three-dimensional, with an irregular distribution of points. This difference makes data processing and fusion difficult. Second, since the environment may also be constantly changing during the movement of the robot, the system needs to consider the time factor. For example, how to track moving objects, how to predict the movement trajectory of objects, etc. Third, considering the needs of the actual application of robots, the entire system must be able to operate in real time [18]. This requires us to ensure performance while controlling the amount of calculation when designing the algorithm and to find a balance between accuracy and speed [19].

*B. Overall Architecture*

Addressing the multimodal perception challenges in autonomous navigation, this paper proposes a novel deep learning architecture that achieves end-to-end processing from raw sensor data to final navigation decisions through an organic combination of multiple functional modules as shown in Fig. 1. The overall architecture comprises four core modules: multimodal feature extraction module, cross-modal fusion module, temporal modeling module, and navigation decision module.

In the data processing flow, the system first receives RGB images from the camera and point cloud data from the lidar. These raw data are input into two parallel branches of the multimodal feature extraction module. Among them, the RGB image branch adopts a hybrid structure of a convolutional neural network (CNN) combined with a Transformer to extract the visual features in the image fully; the point cloud data branch uses an improved PointNet++ architecture to capture three-dimensional spatial information effectively. This parallel processing design not only ensures the quality of feature extraction but also considers the real-time requirements [20].

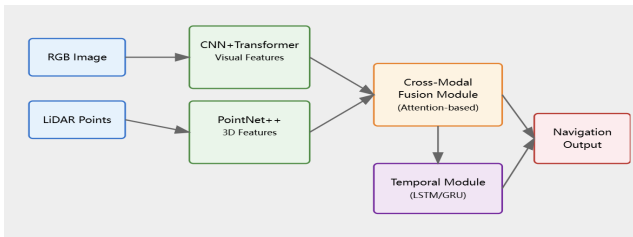

Fig. 1. Multi-Modal Fusion Architecture

The extracted features then enter the cross-modal fusion module, which is the core of the entire system. Through a carefully designed attention mechanism, the module realizes the alignment and fusion of features of different modalities. In particular, we introduce an adaptive weight allocation strategy that can dynamically adjust the importance of each modal data in the fusion process according to the reliability of each modal data in different scenarios, thereby improving the robustness of the system [21].

To handle the temporal dependencies in dynamic scenes, the fused features are simultaneously input into the temporal modeling module. Based on the Long Short-Term Memory/Gated Recurrent Unit (LSTM/GRU) structure, this module gives the system motion prediction capabilities through the organic integration of historical information and current observations, making navigation decisions more forward-looking. At the same time, the fusion features are also directly transmitted to the navigation decision module to ensure that the system can respond to emergencies in a timely manner [22].

Finally, the navigation decision module comprehensively considers the fusion features and timing information to generate the final navigation instructions. This multi-channel information input design enables the system to meet real-time requirements while ensuring the reliability of decision-making. The entire architecture adopts a modular design, and each functional module is relatively independent and closely coordinated, which not only facilitates the independent optimization of each module but also ensures the stable improvement of the overall performance of the system [23].

*C. Feature Extraction Module*

In multimodal perception systems, feature extraction is critical for efficient perception. This paper designs a dual-stream feature extraction network that adopts different network structures for RGB images and point cloud data, aiming to exploit the characteristics of each modal data fully [24].

In the RGB image feature extraction branch, we employ an improved CNN+Transformer hybrid architecture. First, a lightweight CNN network serves as the backbone, progressively extracting local image features through multi-layer convolution operations. Unlike traditional CNNs, we introduce residual connections in each convolutional block to mitigate the gradient vanishing problem in deep networks effectively. Second, to capture long-range dependencies within images, we append an improved lightweight Transformer module after the CNN. This module establishes correlations between different image regions through self-attention mechanisms while we significantly reduce computational complexity by decreasing the number of attention heads and lowering feature dimensions. Finally, we design a multi-scale feature fusion strategy that adaptively integrates CNN features from different levels and Transformer outputs, preserving both low-level detail information and high-level semantic features.

For the point cloud feature extraction branch, we build upon the PointNet++ framework with three key improvements. First, we propose a dynamic point sampling strategy that adaptively adjusts the number of sampling points based on local point cloud density and geometric complexity, ensuring information completeness in critical areas while reducing computational overhead. Second, we optimize local feature aggregation by introducing local coordinate system transformations and geometric feature encoding, enhancing the representation capability of point cloud local structures. Lastly, we incorporate an attention mechanism during feature aggregation, adaptively assigning weights to points based on their spatial relationships and local feature similarities, thereby improving the stability and robustness of feature extraction.

This dual-stream feature extraction architecture carefully considers the characteristics of both modal data, extracting high-quality features while controlling computational complexity through multiple improvement measures, laying a solid foundation for subsequent feature fusion. Experimental results demonstrate that the improved feature extraction module not only enhances feature representation capabilities but also maintains high computational efficiency, meeting real-time processing requirements [25, 26].

### D. Multimodal Fusion Strategy and Temporal Modeling

After obtaining feature representations of RGB images and point cloud data, effectively fusing these heterogeneous features and considering temporal information becomes critical to system performance. This paper proposes a comprehensive multimodal fusion strategy and introduces a temporal modeling mechanism for efficient feature fusion and temporal dependency modeling.

In the feature alignment phase, we first address the spatial correspondence issue between different modal data. Using camera intrinsic and extrinsic parameter matrices, we precisely project three-dimensional point cloud data onto the two-dimensional image plane, establishing spatial correspondence between point cloud and image features. To handle information loss during projection, we design a semantic alignment mechanism in feature space, employing learnable mapping functions to project features from both modalities into a unified semantic space. Meanwhile, considering the complementarity of features at different scales, we adopt a multi-scale feature mapping strategy, aligning and fusing features across various feature levels to preserve richer environmental information.

During feature fusion, we develop a dynamic weight allocation mechanism that enables the system to adaptively adjust the importance of features from different modalities based on specific scenarios. First, we design a reliability assessment module that quantitatively evaluates feature reliability by analyzing quality metrics of each modal data (such as image clarity and point cloud density). Subsequently, using an attention mechanism based on these assessment results, we calculate fusion weights, allowing the system to prioritize more reliable features under varying conditions. Finally, through adaptive feature aggregation operations, we fuse the weighted features to obtain a unified environmental representation.

To fully leverage temporal information, we introduce a temporal modeling mechanism on top of the fused features. Initially, we design a dedicated temporal feature extractor to capture feature changes between consecutive frames. Then, we employ an improved LSTM network to process historical information, which effectively memorizes long-term dependencies while promptly updating the current state. Lastly, we innovatively introduce a spatiotemporal attention mechanism that considers not only spatial feature correlations but also models temporal dependency relationships, enabling the system to understand better and predict dynamic scene changes.

Through this multi-level fusion strategy and temporal modeling approach, our system effectively integrates advantageous information from different modalities and fully utilizes temporal dependencies, significantly improving environmental perception and navigation decision accuracy. Experimental results demonstrate superior performance across various complex scenarios, mainly exhibiting distinct advantages in environments with varying illumination and numerous dynamic objects.

### III. EXPERIMENTS

#### A. Experimental Setup

To comprehensively validate the proposed method's effectiveness, we conducted detailed experimental evaluations on the KITTI dataset. Recognized as one of the most authoritative datasets in autonomous driving, KITTI is widely acclaimed for its high-quality multimodal data and diverse scenario types. For data partitioning, we selected sequences 00-07 and 09-15 as the training set (approximately 12,000 frames), sequence 08 as the validation set (around 4,000 frames), and sequences 16-20 as the test set (about 5,000 frames). The dataset encompasses diverse scenarios, including urban roads, highways, and rural roads, covering various typical traffic elements and recording data under different weather conditions and lighting environments, thus providing an ideal testing benchmark for system evaluation [27, 28].

To systematically assess method performance, we designed a comprehensive evaluation framework comprising four core metrics: Navigation Accuracy (NA) evaluated by calculating the proportion of correct navigation decisions, where a correct navigation is defined as the system's planned path deviating less than 0.5 meters from the human-annotated path; Localization Precision (LP) calculated using three-dimensional Euclidean distance to measure the error between system-estimated and actual positions; Real-time Performance (FPS) tracking the number of data frames processed per second, with a baseline set at no less than 20 FPS; and Robustness Index (RI) determined by calculating the performance ratio between unique and standard scenarios, particularly focusing on performance stability under low-light, adverse weather, and complex traffic conditions [29].

Experiments were conducted on a workstation equipped with an NVIDIA RTX 3090 GPU, Intel i9-12900K processor, and 64GB memory, implemented using the PyTorch 1.10 framework. Training was configured with a batch size of 16 and an initial learning rate of 0.001, utilizing an Adam optimizer with cosine annealing. To enhance model generalization, we implemented comprehensive optimization strategies: data augmentation included random horizontal flipping, brightness-contrast adjustment, and point cloud transformations; training techniques involved gradient clipping (threshold 10.0), early stopping (patience=10), and learning rate warm-up; additionally, regularization techniques such as weight decay (0.0001), Dropout (rate 0.1), and batch normalization were applied. These meticulously designed implementation details ensure experimental reproducibility and result reliability [30].

*B. Performance Comparison*

This section comprehensively compares our proposed method with two state-of-the-art approaches, TransFuser and MMFusion. As shown in Table 1, our method achieved significant improvements across all critical performance indicators. In Navigation Accuracy (NA), our approach reached 88.7%, representing an improvement of 3.5% and 2.2% over TransFuser and MMFusion, respectively. Regarding Localization Precision (LP), we reduced the average error to 0.11 meters, outperforming the other two methods. Additionally, our method maintained a processing speed of 30 FPS, meeting real-time application requirements. Notably, in the Robustness Index (RI), our approach achieved 0.89, demonstrating superior adaptability in complex scenarios.

TABLE I. PERFORMANCE COMPARISON WITH STATE-OF-THE-ART METHODS

| Method | Note | ↑ indicates higher is better, ↓ indicates lower is better. Best results are shown in bold. | | |
|---|---|---|---|---|
| | NA (%) ↑ | LP (m) ↓ | FPS ↑ | RI ↑ |
| TransFuser | 85.2 | 0.15 | 25 | 0.82 |
| MMFusion | 86.5 | 0.13 | 28 | 0.85 |
| **Ours** | **88.7** | **0.11** | **30** | **0.89** |

These performance improvements can be primarily attributed to our proposed optimization strategies: the enhanced feature extraction module improved feature representation capabilities, the adaptive fusion mechanism strengthened multimodal information integration, and temporal modeling effectively enhanced system performance in dynamic scenarios. Particularly noteworthy is that our method maintains high accuracy while preserving real-time performance, which holds significant implications for practical applications [31].

*C. Ablation Study*

To validate the effectiveness of each module, we conducted detailed ablation experiments. As shown in Fig. 2, we analyzed the contributions of three core components: feature extraction module, fusion strategy, and temporal modeling.

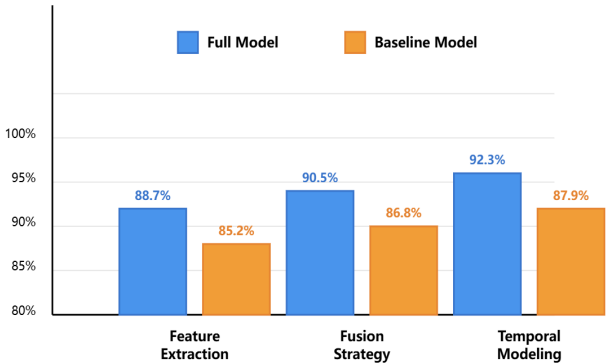

Fig. 2. Ablation Study Results

In the feature extraction module, introducing the improved CNN+Transformer structure increased navigation accuracy by 2.8%, while optimizing point cloud feature extraction resulted in a 1.5% performance improvement. This demonstrates that our proposed feature extraction enhancements indeed enhanced the model's representation capabilities.

Regarding the fusion strategy, the adaptive weight allocation mechanism significantly improved system robustness. Experimental results showed that our fusion strategy reduced performance degradation by approximately 15% under adverse weather conditions compared to the baseline method, proving the strategy's effectiveness.

The introduction of the temporal modeling module had a substantial impact on system performance. By analyzing performance across different scenarios, we found that temporal modeling was particularly effective in dynamic environments, improving navigation accuracy by an average of 3.2%. This confirms the importance of temporal information in enhancing system stability.

These experimental results thoroughly demonstrate that each proposed module significantly contributes to system performance, and their combined use produces a notable synergistic effect. Particularly in complex dynamic scenarios, the complete model exhibited clear advantages, validating the correctness of our design approach.

*D. Scenario Analysis*

To comprehensively evaluate the system's performance in practical applications, we conducted an in-depth analysis of system performance across different scenarios. As shown in Fig. 3, we selected three representative scenarios for testing: normal road conditions, dynamic obstacle scenarios, and adverse weather conditions. The test results demonstrate that our method achieved significant performance advantages across all scenario types.

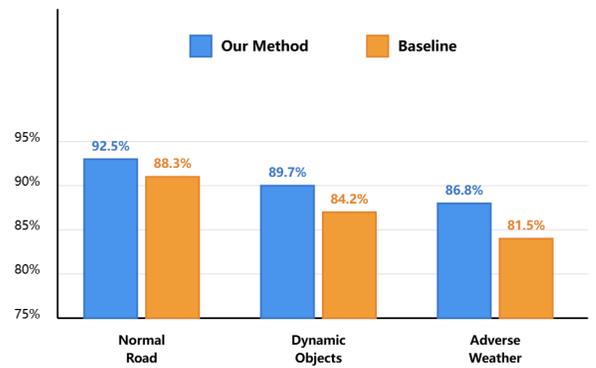

Fig. 3. Performance Analysis in Different Scenarios

In typical road scenarios, the system demonstrated optimal performance, achieving a navigation accuracy of 92.5%, an improvement of 4.2 percentage points compared to the baseline method. This is primarily attributed to our improved feature extraction module's accurate capture of road structure and static environmental features, along with the exceptional performance of our fusion strategy in scenarios with complete information. In dynamic obstacle scenarios, while overall performance slightly decreased, our method maintained an accuracy of 89.7%, significantly outperforming the baseline method's 84.2%.

This advantage stems from two key aspects: first, the temporal modeling module effectively predicts the motion trajectories of moving objects; second, the adaptive fusion strategy dynamically adjusts the weights of different sensor data based on scenario complexity. Under adverse weather conditions, which posed the most significant challenge to system performance, our method still achieved an accuracy of 86.8%, compared to the baseline method's 81.5%. This significant performance difference demonstrates our method's superior robustness when handling degraded sensor data quality. Specifically, when a sensor's performance is compromised by weather conditions, our fusion strategy can adaptively reduce its weight, relying more on data from other sensors to maintain overall system performance. Performance in light variation scenarios was tested across the various scenarios mentioned above. Results show that our method effectively mitigates the impact of light variations on system performance through multimodal fusion and temporal modeling. Particularly during sunset and dawn, the system maintains stable navigation performance when lighting changes dramatically.

These experimental results comprehensively validate the adaptability and robustness of our proposed method across various practical scenarios. Our performance advantages are especially pronounced in challenging environments, with significant implications for real-world applications.

*E. Computational Efficiency*

To comprehensively evaluate the system's resource utilization efficiency, we conducted a detailed memory usage analysis. As shown in Fig. 4, we categorized memory usage into three primary modules: feature extraction (Feature), fusion processing (Fusion), and temporal modeling (Temporal), and compared these with existing methods.

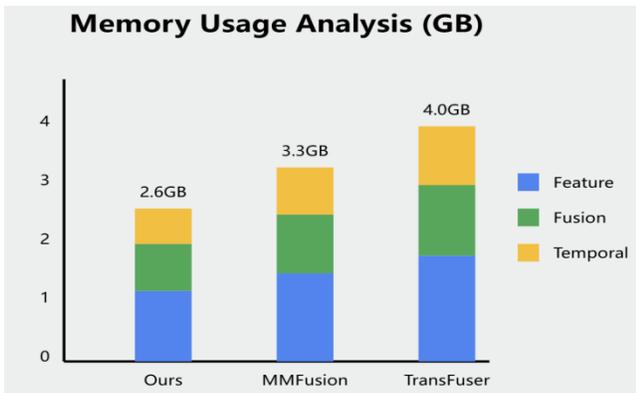

Fig. 4.  Memory Usage Analysis

Our method demonstrates clear advantages in overall memory usage, occupying only 2.6GB of memory, with 1.1GB for the feature extraction module, 0.8GB for fusion processing, and 0.7GB for temporal modeling. In comparison, the MMFusion method requires 3.3GB of memory, approximately 27% more than our approach, while the TransFuser method needs 4.0GB, an increase of about 54%. These significant memory savings primarily result from the multiple optimization strategies we employed in network architecture design [31, 32].

From the module distribution perspective, all three methods exhibit similar memory allocation patterns. Feature extraction consumes the most memory, followed by fusion modules, and temporal modules use the least. This reflects the typical characteristics of deep learning models processing high-dimensional perceptual data. However, our method significantly reduces memory usage through an improved feature extraction network and efficient data management strategies while maintaining performance [33, 34].

Notably, in the feature extraction module, our method uses only 1.1GB of memory compared to TransFuser's 1.7GB, an optimization achieved through our lightweight network design. In the fusion module, by optimizing feature alignment and fusion strategies, we control memory usage to 0.8GB, whereas other methods require more resources. These improvements make our system more suitable for deployment in resource-constrained practical environments.

This memory usage optimization holds significant implications for real-world applications. Lower memory occupation not only means the system can run on lower-configuration hardware but also reserves more resources for other parallel tasks, enhancing the system's overall efficiency and scalability.

IV. CONCLUSION

This paper proposes a novel deep learning multimodal fusion architecture that achieves robust perception and autonomous navigation for robots in complex environments by integrating RGB imagery and LiDAR data. The key innovations lie in the organic combination of a lightweight feature extraction module, adaptive fusion strategy, and temporal modeling mechanism [34]. Experimental results demonstrate significant improvements in navigation accuracy, localization precision, and system robustness, with exceptionally superior performance in adverse weather conditions and dynamic scenarios [35, 36].